\documentclass{article}

\usepackage{amsmath}

\usepackage[final, nonatbib]{neurips_2024}

\usepackage[utf8]{inputenc} 
\usepackage[T1]{fontenc}    
\usepackage{hyperref}       
\usepackage{url}            
\usepackage{booktabs}       
\usepackage{amsfonts}       
\usepackage{nicefrac}       
\usepackage{microtype}      
\usepackage{lipsum}
\usepackage{fancyhdr}       
\usepackage{graphicx}       
\usepackage{float}
\usepackage{enumitem}
\graphicspath{{media/}}

\title{Articulated Animal AI: An Environment for Animal-like Cognition in a Limbed Agent}

\author{
  Jeremy Lucas \\
  McGill University\\
  \texttt{jeremy.lucas@mail.mcgill.ca} \\
   \And
  Isabeau Prémont-Schwarz \\
  Monreal\\
  \texttt{isabeau.premont-schwarz@mail.mcgill.ca} \\
}

\begin{document}
\maketitle

\begin{abstract}
This paper presents the Articulated Animal AI Environment for Animal Cognition, an enhanced version of the previous AnimalAI Environment. Key improvements include the addition of agent limbs, enabling more complex behaviors and interactions with the environment that closely resemble real animal movements. The testbench features an integrated curriculum training sequence and evaluation tools, eliminating the need for users to develop their own training programs. Additionally, the tests and training procedures are randomized, which will improve the agent's generalization capabilities. These advancements significantly expand upon the original AnimalAI framework and will be used to evaluate agents on various aspects of animal cognition.
\end{abstract}

\section{Introduction}
The field of artificial intelligence has seen the development of numerous frameworks designed to test and evaluate models. Each new framework or dataset has often spurred on signmificant progress as different research teams compete to achieve the state the of the art on that specific benchmark. This paper specifically addresses AI frameworks that focus on tasks modeled after animal cognition. Despite significant advancements, many animals still outperform even the most advanced AI systems in various cognitive tasks. A notable framework within this domain is the AnimalAI Environment, which features a simplistic setup with a single spherical agent capable of only basic movements, such as moving forward or backward, and rotating left or right. This environment offers a set of simple building blocks—such as blocks, half-cylinders, spherical pieces of food, ramps, and walls—within a built-in arena, allowing researchers to create their own cognitive tasks for the agent.

However, the AnimalAI Environment presents several limitations. The broad scope of the environment adds complexity to the task of training AI models, as researchers must first design effective tests. Additionally, the environment lacks generalization capabilities; object positions and rotations must be manually configured, leading to the potential for overfitting due to the absence of randomization in tests. Moreover, the agent itself is rudimentary and unrealistic, restricting the exploration of cognitive abilities related to movement. The Articulated Animal AI Environment has been developed to address these shortcomings, offering a more sophisticated and versatile platform for evaluating AI in the context of animal cognition tasks.

\section{The Influence from Animal AI}
The AnimalAI test environment serves as a versatile platform that enables scientists and researchers to design cognitive assessments for a simple agent using modular building blocks. Upon the publication of their paper, the AnimalAI team introduced a competition aimed at developing a highly generalizable agent capable of performing across a diverse array of unknown tests. This competition presented researchers with the challenge of generating a sufficient number of tests to train such an agent. The tests devised for the competition by the AnimalAI team were grounded in ten distinct categories of animal cognition. The results of the competition were disappointing, with even the best AI models performing significantly worse than a human child on every task apart from food retrieval, internal modeling, and numerosity. The meaning and definitions of these categories are described in the next section. Although the competition outcomes were underwhelming, the challenge itself was noteworthy.\cite{animalai2019} \cite{animalai2023}

\section{Animal AI’s Cognition Benchmarks}
We utilized the cognitive categories employed in the AnimalAI competition to develop our benchmark platform. These ten categories are as follows:
\begin{enumerate} [label=L\arabic*.] 
    \item Basic food retrieval: This category tests the agent’s ability to reliably retrieve food in the presence of only food items. It is necessary to make sure agents have the same motivation as animals for subsequent tests.
    \item Preferences: This category tests an agent’s ability to choose the most rewarding course of action. Tests are designed to be unambiguous as to the correct course of action based on the rewards in our environment.
    \item Obstacles: This category contains objects that might impede the agent’s navigation. To succeed, the agent will have to explore its environment, a key component of animal behavior.
    \item Avoidance: This category identifies an agent’s ability to detect and avoid negative stimuli, which is critical for biological organisms and important for subsequent tests.
    \item Spatial Reasoning: This category tests an agent’s ability to understand the spatial affordances of its environment, including knowledge of simple physics and memory of previously visited locations.
    \item Robustness: This category includes variations of the environment that look superficially different, but for which affordances and solutions to problems remain the same.
    \item Internal Models: In these tests, the lights may turn off, and the agent must remember the layout of the environment to navigate in the dark.
    \item Object Permanence and Working Memory: This category checks whether the agent understands that objects persist even when they are out of sight, as they do in the real world and in our environment.
    \item Numerosity: This category tests the agent’s to judge which area has the most food, as it only gets to choose one area.
    \item Causal Reasoning and Object Affordances: This category tests the agent’s ability to use objects to reach it's goal.
\end{enumerate}

\section{Additional Categories}
In addition to the 10 categories listed above, we introduced two additional categories, L0 - Initial Food Contact and L11 - Body Awareness, to address specific challenges posed by the multi-limbed agent within our environment. L0 was created to provide a simpler framework for the agent to learn to move toward food, recognizing the necessity for a foundational category focused on basic movement. L11 was developed in response to the absence of limb-related cognitive tests in the original AnimalAI competition, which featured a spherical, limbless agent. This new category was essential for evaluating the agent's understanding and coordination of its limbs.

\section{Environment Design}
\subsection{The Agent}
\begin{figure}[H]
    \centering
    \includegraphics[width=0.25\textwidth]{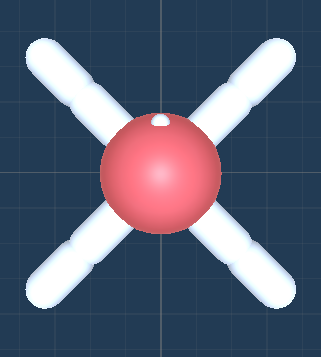}
    \caption{The Top Down view of the agent.}
    \label{fig:agent}
\end{figure}

\textbf{Description:}

The agent is a soft body constructed using configurable joints. It has four thighs and four legs, totaling eight joints. Each joint can rotate in the x direction (up and down) between -90 and 90 degrees and in the z direction (sideways) between -45 and 45 degrees. The head has a mass of 0.5, while each thigh and leg have a mass of 1. The lighter weight of the head helps the agent walk more effectively, preventing it from dragging its head on the ground.

\subsection{Observational Parameters}
The vision space for the agent offers a choice between customizing a camera or using raycasts. By default, the agent's joint rotations are normalized within the range of [-1, 1] and used as observations.

\subsubsection{Camera}
The camera extends from the agent’s eye (indicated by a white dot) and can be set to grayscale. It supports resolutions ranging from 8x8 to 512x512 pixels.

\subsubsection{Raycast}
Raycasts also emanate from the agent’s eye. The viewing angle and number of rays can be customized. The viewing angle ranges from 5 to 180 degrees, encompassing both the left and right sides of the agent. A viewing angle of 180 degrees covers the full 360-degree surroundings of the agent, while a 90-degree viewing angle covers the front half. The number of rays, ranging from 1 to 20, determines the number of rays emitted in each direction. There is always a ray pointing directly forward.

\subsection{Action Parameters}
\subsubsection{Joint Rotation}
Selecting joint rotation enables the agent to move its joints by setting a target rotation property. The joints are propelled to reach the target rotation through motor control.

\subsubsection{Joint Velocity}
Selecting joint velocity allows the agent to move its joints by setting a target angular velocity property. The joints are propelled by adding the specified angular velocity.

\subsection{Rewards}
There are four distinct types of rewards in these environments:
\begin{itemize}
    \item \textbf{Food Pieces:} Green food pieces provide rewards equal to their scale or size. Yellow food pieces provide rewards equal to half of their scale or size.
    \item \textbf{Wall Collisions:} Agents receive a -1 reward when they collide with walls.
    \item \textbf{Training Facilitation:} Agents receive two rewards each timestep, both equal to -(0.5)/maxsteps. The first reward is received only if the head is on the ground, encouraging the agent to walk rather than drag its head. The second reward is given every timestep, incentivizing the agent to explore and make decisions.
\end{itemize}

\subsection{Other Parameters}
\textbf{Maxsteps:} This parameter defines the number of steps the agent has per episode. It is recommended to set this value between 1000 and 5000 steps.

\subsection{The Environment}
\subsubsection{Parameters}
\textbf{Difficulty:} The difficulty parameter ranges from 0 to 10, increasing the challenge of the current level when generated. The specific aspects that become more difficult vary for each level and are described in the test bench.

\textbf{Seed:} The seed parameter allows for the saving of certain configurations for comparative testing. Setting a unique seed ensures that the level will generate and behave consistently every time.

\subsection{The Test Bench}
\textbf{L0:} L0 is the initial test designed to help the agent begin learning to approach and consume food. In this stage, the food is placed directly in front of the agent. As the difficulty level increases, the food is positioned farther away, becomes smaller, and varies in its horizontal placement relative to the agent.
\begin{figure}[H]
    \centering
    \includegraphics[width=0.25\textwidth]{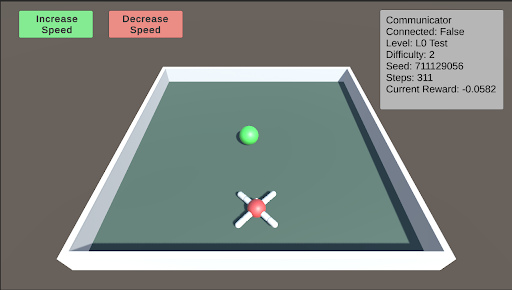}
    \label{fig:agent}
\end{figure}

\textbf{L1: Basic Food Retrieval} \\
L1 serves as a benchmark to test the agent's understanding of the food. The agent must recognize that yellow food pieces do not end the episode, while green food pieces do. As the difficulty increases, more pieces of food are distributed throughout the arena, ranging from 1 to 5 pieces. At lower difficulty levels, the food is mostly stationary, but it begins to move as the difficulty increases.
\begin{figure}[H]
    \centering
    \includegraphics[width=0.25\textwidth]{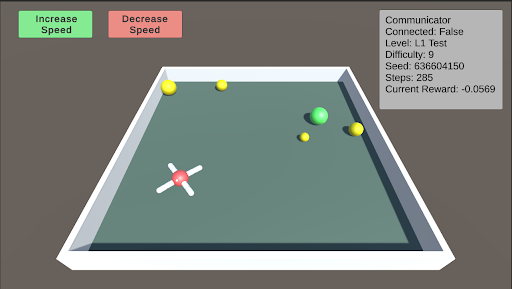}
    \label{fig:agent}
\end{figure}

\textbf{L2: Preferences - Y-Maze} \\
L2 tests the agent’s ability to make choices to obtain better or more food. The Y-maze consists of two paths. When the difficulty is below 5, there will be only one piece of green food at one of the two paths. When the difficulty is above 5, there is an additional smaller piece of green food in the other path. The agent must learn to choose the optimal path.
\begin{figure}[H]
    \centering
    \includegraphics[width=0.25\textwidth]{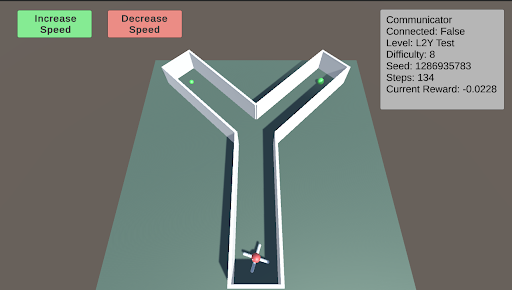}
    \label{fig:agent}
\end{figure}

\textbf{L2: Preferences - Delayed Gratification} \\
This test consists of a small piece of green food in front of the agent and a larger piece of yellow food on top of a pillar that is slowly descending. The agent must learn to wait for the yellow food to come down before consuming the green food. As the difficulty increases, the time it takes for the pillar to descend also increases.
\begin{figure}[H]
    \centering
    \includegraphics[width=0.25\textwidth]{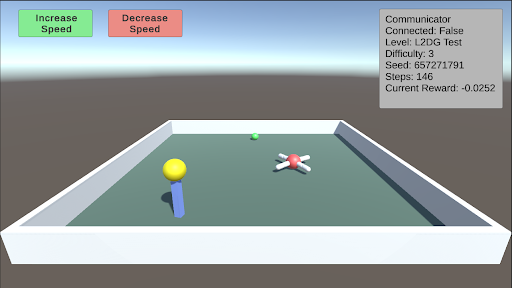}
    \label{fig:agent}
\end{figure}

\textbf{L3: Obstacles} \\
This test involves a single green piece of food placed in a random position and covered by a randomly sized transparent wall. Additionally, other walls will spawn in random orientations as obstacles. As the difficulty increases, both the size of the walls and the number of additional walls increase.
\begin{figure}[H]
    \centering
    \includegraphics[width=0.25\textwidth]{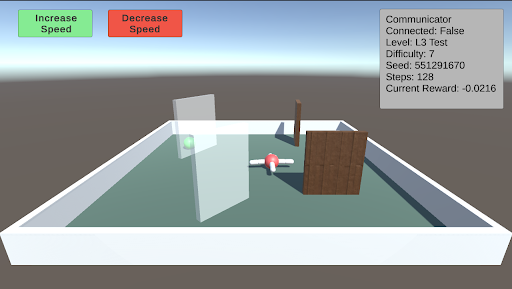}
    \label{fig:agent}
\end{figure}

\textbf{L4: Avoidance} \\
This test involves creating holes in the ground to serve as a negative stimulus. While the previous AnimalAI paper used negative reward floor pads, holes provide a more realistic challenge. As the difficulty increases, the holes become larger. The test starts with two holes, and when the difficulty exceeds 5, a third hole is introduced.
\begin{figure}[H]
    \centering
    \includegraphics[width=0.25\textwidth]{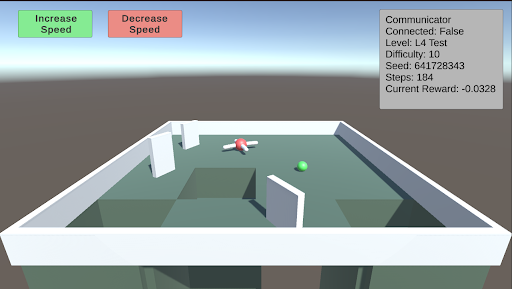}
    \label{fig:agent}
\end{figure}

\textbf{L5: Spatial Reasoning} \\
This test evaluates the agent's ability to navigate a maze to find a piece of food hidden behind a wall in a random sector. Additional walls are placed to obstruct the agent's path. As the difficulty increases, the number and size of the walls increase, and the maze itself becomes larger and more complex.
\begin{figure}[H]
    \centering
    \includegraphics[width=0.25\textwidth]{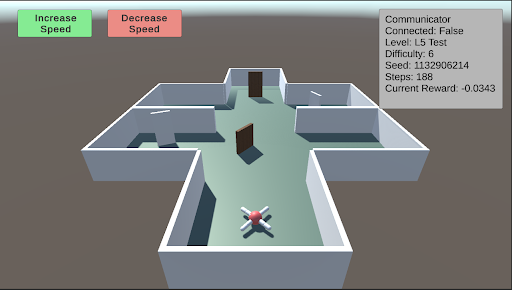}
    \label{fig:agent}
\end{figure}

\textbf{L6: Robustness} \\
Robustness is inherently present in all these tests due to their randomized nature. However, enabling robustness specifically will result in a random test from L0 to L11 being selected, and the colors of the test objects changing to one of five randomly generated colors.
\begin{figure}[H]
    \centering
    \includegraphics[width=0.25\textwidth]{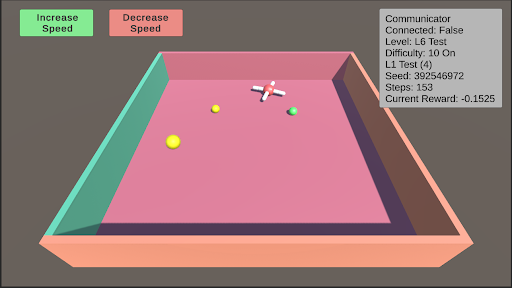}
    \label{fig:agent}
\end{figure}

\textbf{L7: Internal Models} \\
This test assesses the agent's internal modeling capabilities using its camera. Blackouts will occur, with the frequency and duration of these blackouts increasing as the difficulty rises. During these blackouts, the agent's camera will render a dark screen. Enabling internal models will select a random test from L0 to L11 and apply the blackouts described above. 
\begin{figure}[H]
    \centering
    \includegraphics[width=0.25\textwidth]{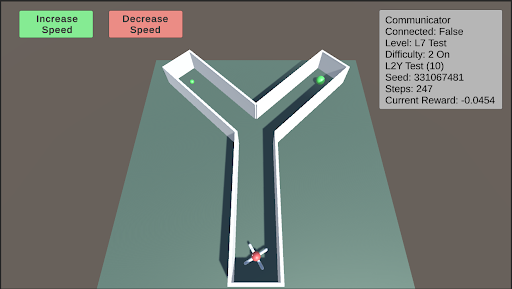}
    \label{fig:agent}
\end{figure}
\textbf{L8: Object Permanence} \\
This test involves a green piece of food that spawns on either side of a wall facing the agent. The food then moves behind the wall. There is a barrier separating the two sides behind the wall, with one side containing the food and the other side containing a hole. The agent must remember that the food still exists behind the wall. As difficulty increases, the speed at which the food moves behind the wall also increases.
\begin{figure}[H]
    \centering
    \includegraphics[width=0.25\textwidth]{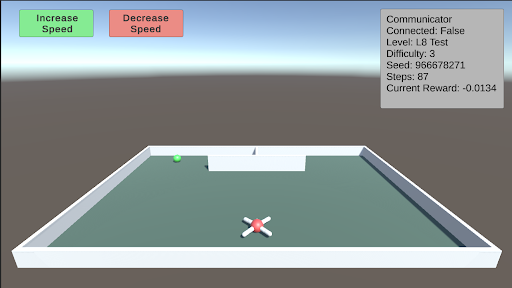}
    \label{fig:agent}
\end{figure}

\textbf{L9: Numerosity} \\
This test requires the agent to choose between different sectors based on the quantity of food present. There are four sectors, each containing a random number of food pieces. The agent must enter a sector and collect all the food within it. Once the agent enters a sector, the opening will close. The agent must correctly choose the sector with the most food. As the difficulty increases, the number of food pieces in each sector increases, making the decision more challenging.
\begin{figure}[H]
    \centering
    \includegraphics[width=0.25\textwidth]{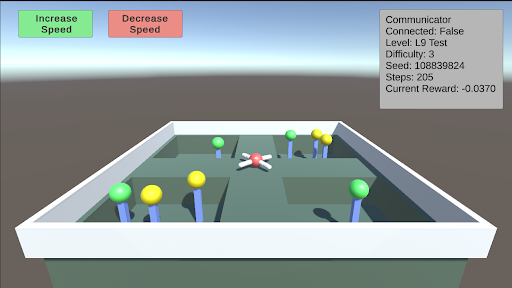}
    \label{fig:agent}
\end{figure}

\textbf{L10: Causal Reasoning} \\
This test evaluates the agent's cognitive ability to understand the environment around it. By pushing over a plank, the agent can walk over a trench to reach a piece of food. However, as the difficulty increases, transparent and opaque walls will appear alongside the plank. The agent must comprehend the cause and effect of its actions, and pick the plank to push over. Additionally, the trench will gradually become wider. 
\begin{figure}[H]
    \centering
    \includegraphics[width=0.25\textwidth]{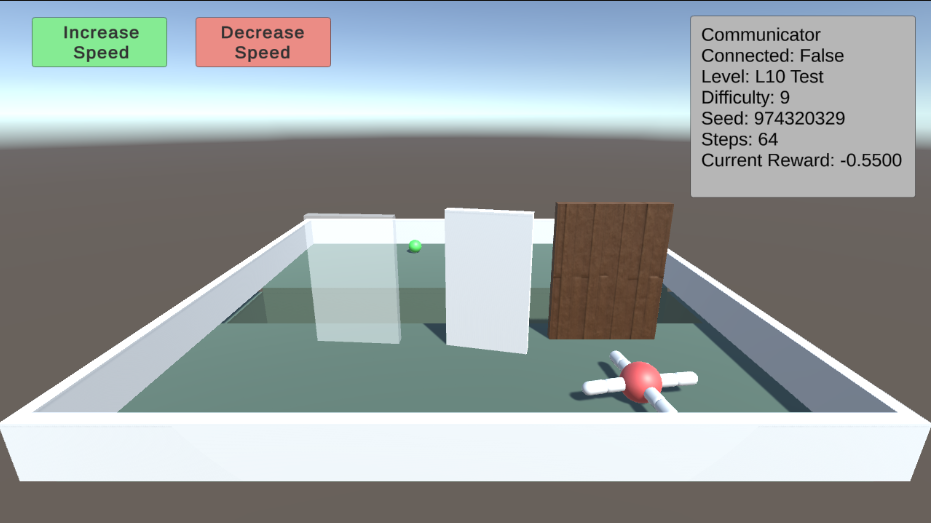}
    \label{fig:agent}
\end{figure}
\textbf{L11: Body Awareness} \\
This test assesses the agent's understanding of its own body. After navigating through the corridors of the Y-maze, Thorndike hut, and a normal maze, the agent has developed a basic understanding of its body. This test further evaluates the agent's ability to function on rough terrain. A single piece of food is placed away from the agent on uneven terrain. As the difficulty increases, the terrain becomes progressively rougher.
\begin{figure}[H]
    \centering
    \includegraphics[width=0.25\textwidth]{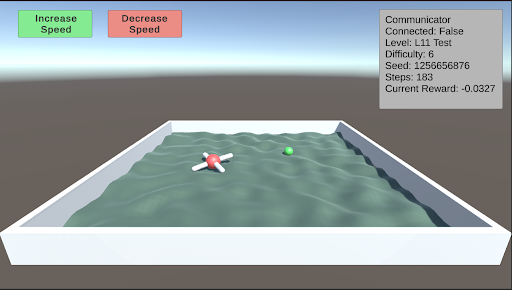}
    \label{fig:agent}
\end{figure}

\section{Curriculum Training}
We provide a curriculum training inspired by \cite{cumulativetrainingarticle} where the agent focuses on tasks were the learning progress (or regression) is fastest. The system operates by organizing the levels and difficulties into a 2D matrix, where each cell represents a specific level at a given difficulty. Upon the completion of an episode, the agent's reward is recorded in the corresponding cell of the matrix. A rolling moving average and variance are calculated for each cell over a period of 10 episodes, and the z-value for each cell is determined using the following equation.

\begin{equation}
    z_i := \frac{\sigma_i}{|\mu_i| + 10^{-7}}
\end{equation}

These z-values are then used to create a probability distribution that guides the selection of the next training level and difficulty, prioritizing levels with higher variance. Thus the agent will focus more on tasks where it's performance is highly variable. The probability of each cell being selected next is determined according to the equation below.

\begin{equation}
   p_i = \frac{z_i + c}{\sum_{j=1}^{n} (z_j + c)}
\end{equation}

This curriculum training procedure effectively promotes learning and facilitates training. As the agent begins to learn and increases its end reward for a task, the rolling variance rises, which in turn increases the z-value and the likelihood of the same level and difficulty being presented again. This creates a feedback loop until the agent either stops learning or completes the current level and difficulty. As the variance decreases, a new level and difficulty pair is presented to the agent for further learning. Additionally, this training combats catastrophic forgetting. If the agent revisits a previously mastered level and difficulty pair but is currently failing, the increased variance will cause this pair to reappear more frequently until the task is relearned. Overall, this curriculum training procedure provides an effective method for training an agent across a diverse range of tasks.

\section{Future Work}

In the initial release of the Articulated Animal AI Environment, our goal is to enable researchers to familiarize themselves with the platform and test their cognition-based models using the provided test bench. A designated set of testing seeds has been reserved within the environment, which will be used for future evaluations of selected candidates or potential competition scenarios.

\section{Conclusion}
The Articulated Animal AI Environment is a robust framework designed to assess the capabilities of a multi-limbed agent in performing animal cognition tasks. Building upon the foundational AnimalAI Environment, the Articulated Animal AI Environment enhances the training process by offering pre-randomized and generalized tests, coupled with a structured curriculum training program and comprehensive evaluation tools. This setup allows researchers to concentrate on refining their models rather than on the time-intensive task of developing effective tests.

\section*{Acknowledgments}
This work was supported in part by McGill University.

\bibliographystyle{unsrt}  
\bibliography{references}  

\begin{thebibliography}{1}

\bibitem{animalai2019}
Matthew Crosby, Benjamin Beyret, Murray Shanahan, Jos{\'e} Hern{\'a}ndez-Orallo, Lucy Cheke, and Marta Halina.
\newblock Animal-ai: A testbed for experiments on autonomous agents.
\newblock {\em arXiv preprint arXiv:1909.07483}, 2019.

\bibitem{animalai2023}
Konstantinos Voudouris, Ibrahim Alhas, Wout Schellaert, Matthew Crosby, Joel Holmes, John Burden, Niharika Chaubey, Niall Donnelly, Matishalin Patel, Marta Halina, José Hernández-Orallo, and Lucy~G. Cheke.
\newblock Animal-ai 3: What's new \& why you should care.
\newblock {\em arXiv preprint arXiv:2312.11414}, 2023.

\bibitem{cumulativetrainingarticle}
Jacqueline Gottlieb, Pierre-Yves Oudeyer, Manuel Lopes, and Adrien Baranes.
\newblock Information seeking, curiosity and attention: computational and neural mechanisms.
\newblock {\em Trends in Cognitive Sciences}, 17(11):585--593, 2013.

\end{thebibliography}

\appendix
\newpage
\section{Appendix}
This appendix contains additional information about how to use the interface. All source code is available online. 

Environment source code: https://github.com/jeremy-lucas-mcgill/Articulated-Animal-AI-Environment

\section*{The Interface}
The interface provides users with the capability to customize their agents by selecting specific actions, vision settings, and general parameters. Users can then choose the levels and difficulty range for training. The interface offers two primary options for initiating the training process: "Start Random," which randomly selects a level and difficulty for each episode, or "Start Curriculum," which begins the curriculum-based training program.
\begin{figure}[H]
    \centering
    \includegraphics[width=0.5\textwidth]{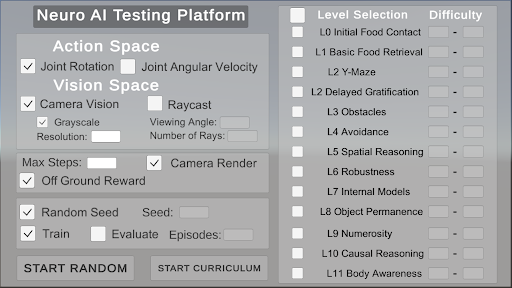}
    \label{fig:agent}
\end{figure}

\end{document}